\documentclass[10pt,twocolumn,letterpaper]{article}

\usepackage{cvpr}  
\usepackage{algorithm}
\usepackage{algpseudocodex}
\definecolor{cvprblue}{rgb}{0.21,0.49,0.74}
\usepackage[pagebackref,breaklinks,colorlinks,citecolor=cvprblue]{hyperref}

\title{Spatial-Temporal Conditional Random Field for Human Trajectory Prediction}

\author{Pengqian Han\\
The University of Auckland\\
Auckland, New Zealand.
\and
Jiamou Liu\\
The University of Auckland\\
Auckland, New Zealand.
\and 
Jialing He\\
Chongqing University\\
Chongqing, China.
\and 
Zeyu Zhang\\
Huazhong Agricultural University\\
Wuhan, China.
\and
Song Yang\\
The University of Auckland\\
Auckland, New Zealand.
\and 
Yanni Tang\\
The University of Auckland\\
Auckland, New Zealand.
}

\begin{document}
\maketitle
\begin{abstract}
Trajectory prediction is of significant importance in computer vision. Accurate pedestrian trajectory prediction benefits autonomous vehicles and robots in planning their motion. Pedestrians' trajectories are greatly influenced by their intentions. Prior studies having introduced various deep learning methods only pay attention to the spatial and temporal information of trajectory, overlooking the explicit intention information. In this study, we introduce a novel model, termed the \textbf{S-T CRF}: \textbf{S}patial-\textbf{T}emporal \textbf{C}onditional \textbf{R}andom \textbf{F}ield, which judiciously incorporates intention information besides spatial and temporal information of trajectory. This model uses a Conditional Random Field (CRF) to generate a representation of future intentions, greatly improving the prediction of subsequent trajectories when combined with spatial-temporal representation. Furthermore, the study innovatively devises a space CRF loss and a time CRF loss, meticulously designed to enhance interaction constraints and temporal dynamics, respectively. Extensive experimental evaluations on dataset ETH/UCY and SDD demonstrate that the proposed method surpasses existing baseline approaches.
\end{abstract}    
\section{Introduction}
\label{sec:intro}
Predicting the future trajectory of pedestrians \cite{sun2022human, yue2022human, bae2022non, yang2023long} is about anticipating the prospective path of moving people by analyzing their past movements. Accurate trajectory prediction is essential for ensuring safe motion planning in autonomous vehicles. It aids in preventing potential collisions and maintaining a reasonable distance between pedestrians and vehicles \cite{poibrenski2021multimodal, sheng2022graph, bhatt2023mpc}. For instance, autonomous vehicles can effectively plan routes and avoid getting too close to pedestrians in urban areas by accurately predicting the trajectory of pedestrians around them. Additionally, trajectory prediction can assist robots in navigating optimal routes within crowded scenarios \cite{scheggi2014cooperative-robotnavigation,rudenko2020human} and enable surveillance systems to detect abnormal behaviors \cite{luber2010people}.

\begin{figure}
    \centering
    \includegraphics[width=0.5\textwidth]{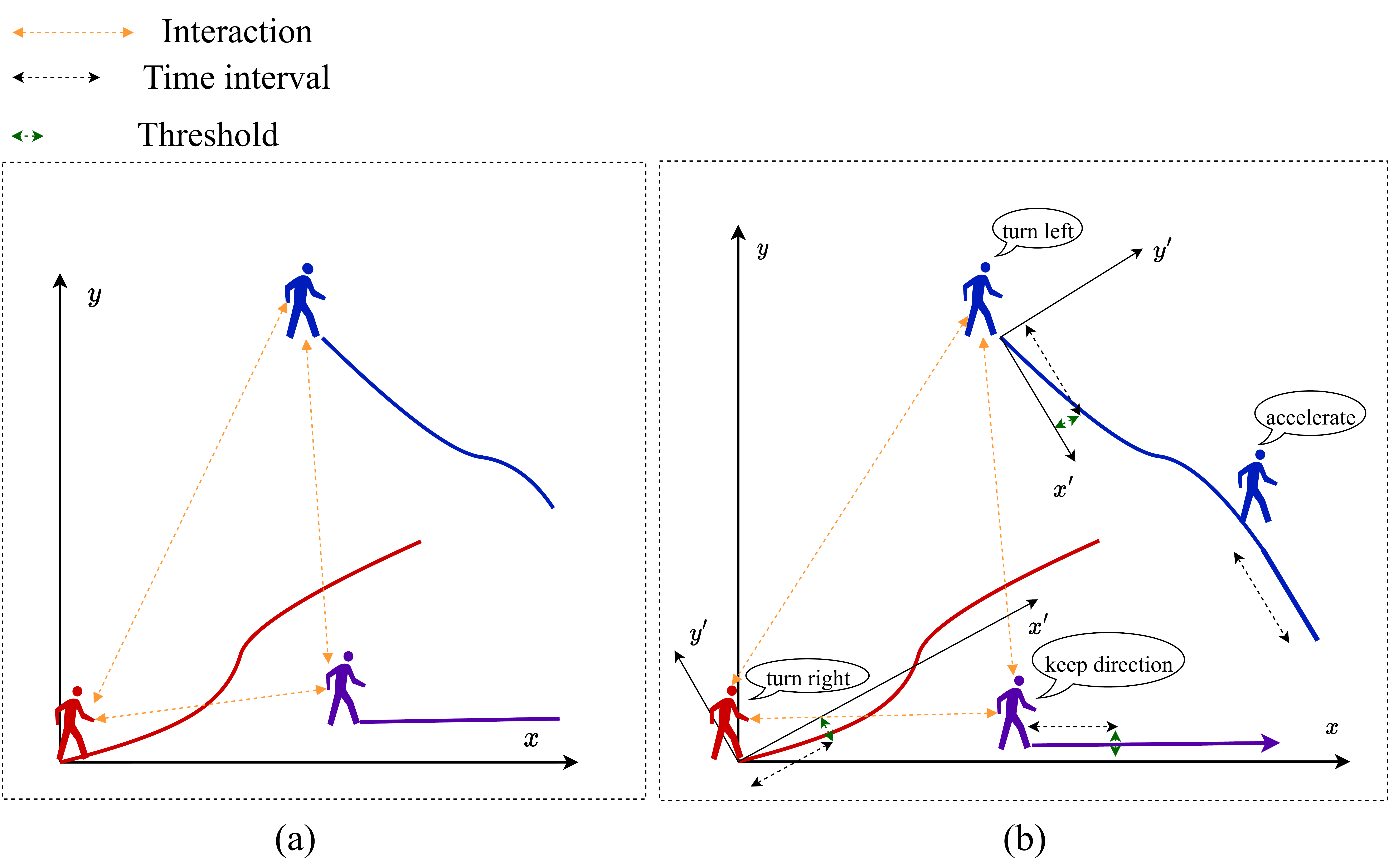}
    \caption{\textbf{Comparison between previous works and S-T CRF:} (a) indicates that previous works solely concentrate on trajectories' spatial and temporal aspects, neglecting the step intention information of individuals in motion. (b) After computing the step intention in the short Time interval, S-T CRF can explicitly capture pedestrians' step intentions as well as the spatial and temporal information in the trajectory data}
    \label{fig:prework_ourwork}
\end{figure}

It is challenging to accurately predict the trajectories of humans due to their properties. Human trajectory data inherently includes three key properties: 1) temporal information, indicating sequential positional changes \cite{cheng2023gatraj}; 2) spatial dimensions, reflecting interaction among pedestrians \cite{huang2023multimodal}; 3) intention information, showing the trend of trajectory change \cite{gomes2022review-intention-aware}. The intention encompasses both lateral intentions (turning left, turning right, keep direction) and longitudinal intentions (accelerating, decelerating, uniform speed). To utilize the intention information, we define an algorithm to compute the step intention as shown in \ref{alg:ma_compute}, which computes the step intention of the pedestrian in a short time interval. 

Although the intention information corresponds to the future trajectory of pedestrians \cite{zhang2021vehicle-intention}, most previous studies \cite{alahiSocialLSTMHuman2016a-sociallstm,chandra2019traphic, song2020pip, li2019grip++,mohamed2020social-social-stgcnn} have focused on only the first two pieces of information. They disregard pedestrians' explicit intentions which is a prevalent feature in pedestrian trajectory datasets. Based on the definition of the step intention of the pedestrian shown in \ref{alg:ma_compute}, we analyze the intention in some datasets. As indicated in Tab. \ref{tab:intention_percent}, the combined percentage of turn right and turn left is approximately 20\% in the ETH \cite{pellegriniYouLlNever2009-eth}/UCY \cite{lernerCrowdsExample2007-ucy} dataset and exceeds 40\% in the SDD \cite{robicquet2016learning-sdd} dataset for lateral intentions. Regarding longitudinal intentions, the combined percentage of acceleration and deceleration surpasses 40\% in both the ETH and SDD datasets. Previous research only focuses on spatial and temporal information, disregarding these intentions of pedestrians such as turning right or left, acceleration, and deceleration.
\begin{table}[]
\centering
\caption{Intentions in ETH/UCY and SDD}
\label{tab:intention_percent}
\begin{tabular}{c|c|c}
\toprule
Intention      & ETH/UCY & SDD     \\
\midrule
turn right     & 10.56\% & 30.34\% \\
turn left      & 8.99\%  & 12.25\% \\
keep direction & 80.45\% & 57.42\% \\
\midrule
acceleration   & 23.87\% & 22.06\% \\
deceleration   & 16.03\% & 17.76\% \\
keep speed     & 60.09\% & 60.18\%\\
\bottomrule
\end{tabular}
\end{table}

While some studies like\cite{deoConvolutionalSocialPooling2018a-cslstm, song2020pip, zhang2021vehicle-intention, hao2020attention-driverintentionrecognition} consider vehicle maneuver as driver turning intention information, they approach the prediction of maneuvers as a basic classification problem. However, in practice, driver intention and vehicle maneuvers are influenced by previous actions. For instance, it is important to acknowledge that the maneuver of a car at time $(t+1)$ is influenced by the maneuver at time $t$, highlighting the complexity of predicting maneuvers. Therefore, it is necessary to model the intentions of people.

Pedestrians' intentions are vital in determining their movement. Their future trajectory is greatly influenced by their intentions, such as turning left or right. Additionally, these intentions are dynamic and can be affected by other pedestrians nearby. Fig. \ref{fig:prework_ourwork} (a) shows that previous studies have developed deep learning models to capture the interaction between pedestrians and temporal information, which represents trajectory over time. However, these studies overlook the fact that pedestrians have intentions like turning left or turning right to move when they change their position. In Fig. \ref{fig:prework_ourwork} (b), our work takes into account not only spatial and temporal information but also the intentions of moving pedestrians. Deep learning models for motion forecasting should intuitively consider pedestrians' intentions to ensure accurate trajectory prediction.




In this study, we introduce S-T CRF, a deep learning model trained in end-to-end manner that can capture pedestrians' intention information. As mentioned above, there are three properties in trajectory. The S-T CRF can capture spatial, temporal, and intention information. The spatial-temporal module is designed to learn the spatial-temporal representation of the observed trajectory. In the model, the intention computer is used to compute the intention label based on the original trajectories based on the lateral threshold and longitudinal threshold. The labels are used as intention tags in time CRF and space CRF. For the time CRF, it outputs the intentions representation during the predicted time. The space CRF loss can enhance the representation of the spatial-temporal module and make the interaction information more explainable. The time CRF loss can capture more temporal information and make the temporal information more explainable. The model is trained in an end-to-end manner. The CRF modules can be added to most of the trajectory prediction models to improve the performance.

We conducted numerous experiments on ETH/UCY and SDD, demonstrating the superior performance of the proposed model. Our results on the SDD dataset are state-of-the-art (SOTA). One of the ablation study indicates that space CRF loss and time CRF loss contribute to a 13.63\% improvement in average ADE and a 12.00\% improvement in average FDE for S-STGCNN's performance. Another ablation study demonstrates that incorporating time CRF loss and space CRF loss improves prediction performance.

The contributions of this paper are threefold: 
\begin{itemize}

    \item We develop a plug-and-play intention computer that computes the intention label from trajectory data, compatible with various trajectory prediction models.
    \item We propose S-T CRF, which combines time CRF loss and space CRF loss to enhance the representation of GCN and CNN. The predicted intention output by time CRF is combined with spatial-temporal representation to improve prediction accuracy.
    \item  We conduct multiple experiments on various datasets, showcasing superior trajectory prediction performance compared to the baseline. The alation study experiments indicate the usefulness of space CRF loss and time CRF loss.
\end{itemize}

\section{Related Work}
\label{sec:related_w}
\subsection{Trajectory prediction}
Trajectory prediction includes vehicle trajectory prediction and pedestrians trajectory prediction. Conventional methods, such as Hidden Markov Models (HMMs), Support Vector Machines (SVMs), and Dynamic Bayesian Networks (DBNs), have been employed to tackle this issue. Nevertheless, these approaches are constrained by short time horizons, typically around 2 seconds, and their predictive capabilities fall short of expectations. Recently, deep learning methods have attracted more and more attention.
Social LSTM \cite{alahiSocialLSTMHuman2016a-sociallstm} applied LSTM and designed a grid-based social pooling technique between LSTM cells to predict the pedestrians' trajectory. Many works like \cite{deoConvolutionalSocialPooling2018a-cslstm, gupta2018social-socialgan,ma2019trafficpredict-apollodataset,song2020pip} follow this direction, using a pooling mechanism to capture the spatial information and then using RNNs like LSTM to capture time information. But this kind of method does not consider different weights assigned to neighbor agents. Afterward, some work \cite{mohamed2020social-social-stgcnn, li2019grip++, sheng2022graph-gstcn, huang2019stgat} choose graph neural networks (GNNs) to encode the space information. The drawback of these methods is that the spatial and temporal information is processed separately, which means missed information on the historical trajectory. 
\subsection{Graph Neural Networks (GNNs)}
GNNs have emerged as a powerful tool for learning graph-structured data, which is non-Euclidean data \cite{zhang2023csg,jin2023survey,zhang2023rsgnn}. GNNs are able to capture complex relationships and interactions within the data \cite{mohamed2020social-social-stgcnn}. They have been widely applied in various domains, including computer vision, natural language processing, and social network analysis \cite{zhou2020graph-gnnsurvey}. In the context of trajectory prediction, GNNs have shown significant promise. Social-STGCNN \cite{mohamed2020social-social-stgcnn} employs a graph to model interactions, designing a kernel function to compute weighted edges between pedestrians and utilizing a GCN layer to capture spatial information. This work's assumption of a fixed number of nodes contradicts the dynamic nature of the pedestrian movement. In GRIP++ \cite{li2019grip++}, edges between nodes (vehicles) are determined by a relative distance threshold, and Graph Convolution layers and 2D temporal convolution layers are alternated to capture spatial and temporal information. GSTCN \cite{sheng2022graph-gstcn} defines a complete graph with weighted edges, reciprocally based on node distances, and uses GCN layers to model vehicle interaction. ReCoG \cite{mo2020recog} constructs a directed graph with a distance threshold of 20 meters, encoding a square image around the target vehicle with a CNN. The directed heterogeneous graph is processed by two-layer GCNs to capture interaction features. RSBG \cite{sun2020recursive-rsbg} employs a Recursive Social Behavior Generator, trained through annotated social data, to output weighted edges for social interaction propagation. SGCN \cite{shi2021sgcn} addresses the issue of superfluous edges in pedestrian interaction by defining a dense undirected graph, processed into a directed graph through asymmetric convolutional networks. Utilizing High Definition (HD) map data, the STGAT method \cite{huang2019stgat} constructs a complete graph for pedestrian interactions, employing Graph Attention Networks (GAT) and LSTM cells to model temporal dependencies. The Social-BiGAT model \cite{kosaraju2019social-social-bigat} leverages a GAT to process pedestrian hidden states and a latent encoder to generate interaction-inclusive noise, enhancing model multimodality. Recognizing the heterogeneous nature of traffic participants, \cite{mo2022multi-heat} introduces Directed Heterogeneous Graphs and HEAT layers to represent complex interactions, concatenating node embedding, edge attributes, and edge types for the attention mechanism.

\subsection{Conditional Random Fields (CRFs)}
CRF is a probabilistic graphical model that can effectively capture sequential data by considering contextual information \cite{lafferty2001conditional}. Additionally, it has the capability to incorporate more labels in the data, thereby enhancing the performance of deep learning models. CRFs find wide applications across domains such as natural language processing, computer vision, and trajectory prediction. In the field of natural language processing, CRFs are commonly employed by adding additional layers to leverage sentence tags \cite{huang2015bidirectional-lstmcrf}. For instance, \cite{krahenbuhl2011efficient-fullycrf} presents an efficient inference algorithm for fully connected CRFs in multi-class image segmentation. This approach utilizes pairwise potentials between pixels within the CRFs model to improve accuracy in segmentation and labeling tasks. Moreover, \cite{wang2018human-lstmcrfdriving} redefines the maneuver decision problem as a sequence labeling problem. In this work, CRFs are combined with LSTM to model relationships between decisions and address transitions among hidden states. This integration aids in smoothing the outputs of LSTM and enables the capturing of sequential patterns inherent in driving decisions like lane changes. The inclusion of a CRF layer within this architecture ensures that decisions are logically related and follow a coherent sequence similar to how human drivers make decisions.
\section{Problem Formulation}
\label{sec:problem_f}

To formalize the problem, let $N$ denote the number of pedestrians in the time sequence. $\mathcal{N}=\left\{1,2,\dots,N\right\}$ denotes the set of pedestrians indices. The observed time spans from $T_1$ to $T_o$, while the predicted time spans from $T_{o+1}$ to $T_p$. $L_o$ and $L_p$ represent the observed and predicted time lengths, respectively. The input includes the historical trajectory of coordinates during $L_o$ time. So, the training set is denoted as follows
\begin{equation}
    \begin{aligned}
            &\mathbf{X} = \left\{\left(x_i^t, y_i^t\right) | t \in \mathcal{T}_o, i \in \mathcal{N} \right\} \in \mathbb{R}^{2\times  L_o\times N}\\
            &\mathbf{Y} = \left\{\left(x_i^t, y_i^t\right) | t \in \mathcal{T}_p, i \in \mathcal{N} \right\} \in \mathbb{R}^{2\times  L_p\times N}
    \end{aligned}
    \label{eq:input_o}
\end{equation}
where $(x_i^t, y_i^t )$ are the $ith $ coordinates of the pedestrian at time $t$. $\mathcal{T}_o=\left\{T_1,T_2,\dots,T_o\right\}$ represents the set of observed time indices, while $\mathcal{T}_p =\left\{T_o+1,T_o+2,\dots,T_p\right\}$ represents the set of predicted time indices.

To use intention information, we compute the intention label from the trajectory data, which will be explained in section \ref{sec:model}. The intention is denoted as follows
\begin{equation}
    \begin{aligned}
            &\mathbf{I_o} = \left\{(\alpha_i^t, \beta_i^t)| t \in \mathcal{T}_o, i \in \mathcal{N}\right\} \in \mathbb{R}^{2\times  L_o\times N} \\
            &\mathbf{I_p} = \left\{(\alpha_i^t, \beta_i^t)| t \in \mathcal{T}_p, i \in \mathcal{N}\right\} \in \mathbb{R}^{2\times  L_p\times N}
    \end{aligned}
\end{equation}
where $(\alpha_i^t, \beta_i^t)$ is the $ith$ pedestrian's latitudinal and longitudinal intention labels at time $t$. These two labels are computed from the original coordinates $\mathbf{I_o},\mathbf{I_p} = g\left(\mathbf{X},\mathbf{Y}\right)$, which will be explained in section \ref{sec:model}. The latitudinal labels $\alpha_i^t$ include turn left (2), turn right (1), and keep direction (0), while the longitudinal labels $\beta_i^t$ include acceleration (2), acceleration (1), and keep-speed (0). During the training phase, $\mathbf{I_o}$ is used to compute the space CRF loss and $\mathbf{I_p}$ is used for the time CRF loss.
The training phase, as depicted in \ref{eq:train}, involves training a function $f$ by minimizing the loss $\mathcal{L}$.
\begin{equation}
    \mathcal{L} = f\left(\mathbf{X},g\left(\mathbf{X},\mathbf{Y}\right)\right)
    \label{eq:train}
\end{equation}

The output of the model is the future trajectory $\mathbf{\hat{Y}}$ and intentions  $\mathbf{\hat{I}_p}$ of $N$ pedestrians during prediction time. 
shown in Eq. \ref{eq:pred} 
\begin{equation}
    \begin{aligned}
    &\mathbf{\hat{Y}} = \left\{\left(x_i^t, y_i^t\right) | t \in \mathcal{T}_p, i \in \mathcal{N} \right\} \in \mathbb{R}^{2\times  L_p\times N }\\
    &\mathbf{\hat{I}_p} = \left\{(\alpha_i^t, \beta_i^t)| t \in \mathcal{T}_p, i \in \mathcal{N}\right\} \in \mathbb{R}^{2\times  L_p\times N}
    \end{aligned}
    \label{eq:pred}
\end{equation}
The objective of the paper shown in Eq. \ref{eq:objective}  is to use the trained model to predict the future trajectory of pedestrians $\mathbf{\hat{Y}}$. 
\begin{equation}
    \mathbf{\hat{Y}} = f\left(\mathbf{X}\right)
    \label{eq:objective}
\end{equation}


\section{The ST-CRF model}
\label{sec:model}

\begin{figure*}
    \centering
    \includegraphics[width=0.8\linewidth]{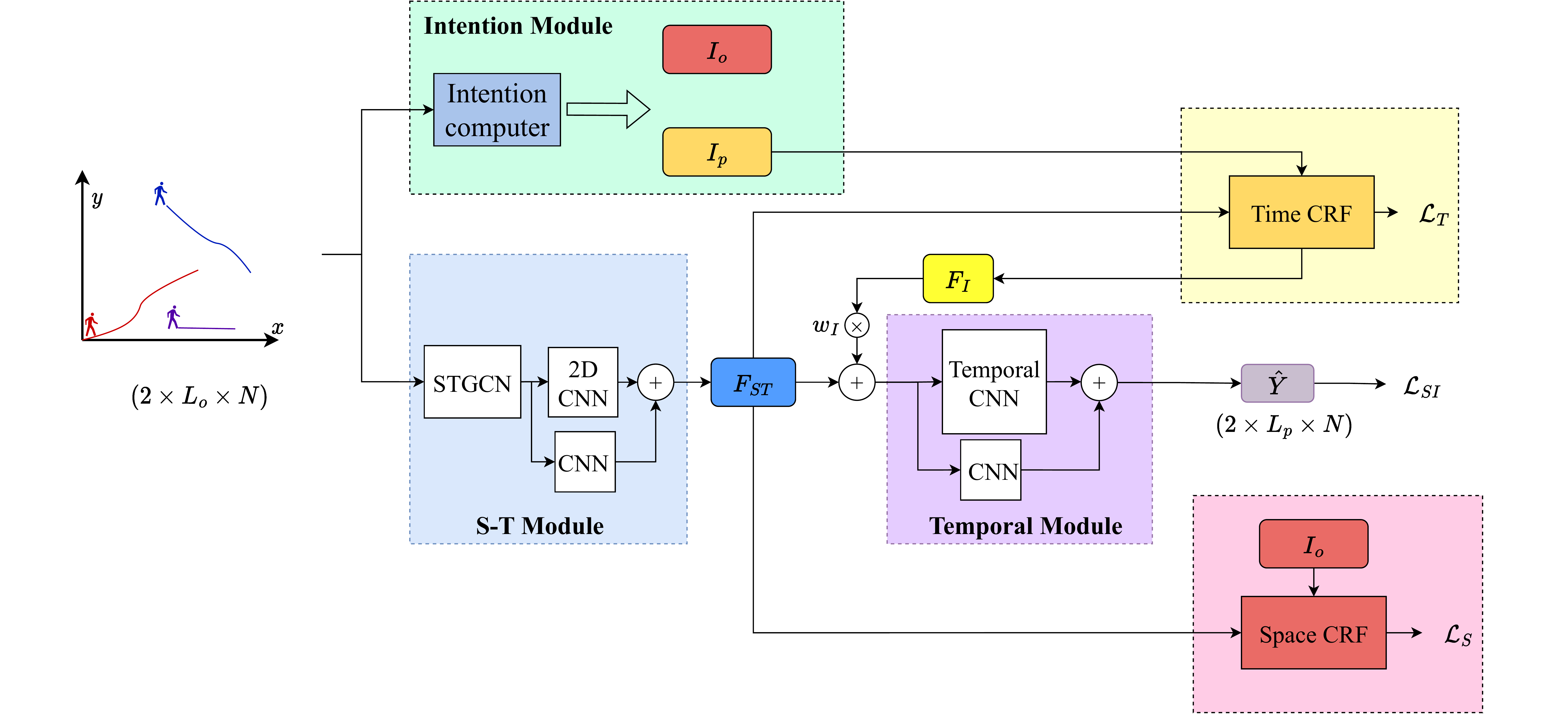}
    \caption{The overall architecture of ST-CRF. The ST-CRF involves feeding the trajectory into the intention computer to obtain the intention label, which is used in both time CRF and space CRF. Simultaneously, the trajectory is inputted into the S-T module to obtain the spatial-temporal representation $F_{ST}$. The $F_{ST}$ is then utilized by both time CRF and space CRF to generate time CRF loss and space CRF loss. Another output of time CRF is the future intention representation $F_I$, which combines with $F_{ST}$ as input for the temporal module, resulting in the output of future trajectory.}
    \label{fig:stcrf_scheme}
\end{figure*}

As depicted in Fig. \ref{fig:stcrf_scheme}, the ST-CRF model consists of five primary modules: intention, Spatial-temporal (S-T), temporal, time CRF, and space CRF. These modules will be discussed in detail in the subsequent sections.

\subsection{Intention module}
The primary objective of the intention block is to calculate the intention labels, which are subsequently supplied to the CRF blocks for the computation of the CRF loss. The intention labels encompass both latitudinal and longitudinal dimensions. The latitudinal labels, delineated as keep direction, left turn, and right turn, are distinctly identified. Specifically, in our implementation, the label `0' denotes keep direction, while `1' and `2' signify turning left and right, respectively. Regarding longitudinal movements, a label of '0' denotes keep speed, `1' represents deceleration, and `2' signifies acceleration.

Intention labels, covering longitudinal and lateral motion, are instrumental in enhancing the performance of deep learning models. These labels encapsulate information regarding motion tendency, which is crucial for future trajectory prediction. The information on motion tendency is encapsulated through the CRF.

The intention label is obtained from each pedestrian's original data, comprising the coordinates of the $x$ and $y$. Since the origin of the original coordinate system is not at the core of the pedestrian, it is necessary to process the original trajectory. After rotating the original coordinates, we establish a new coordinate system. As depicted in Fig. \ref{fig:prework_ourwork} (b), the tangent line at the initial time point of the observed trajectory is defined as the new $x$ axis, denoted as $x'$. The orthogonal direction is defined as the lateral direction, denoted as $y'$. The $x'$, $y'$ axes are exclusively utilized to compute the intention label. Subsequently, the computation of the intention label is performed. Alg. \ref{alg:ma_compute} demonstrates the computation process for lateral and longitudinal labels.

\begin{figure}
    \centering
    \includegraphics[width=0.8\linewidth]{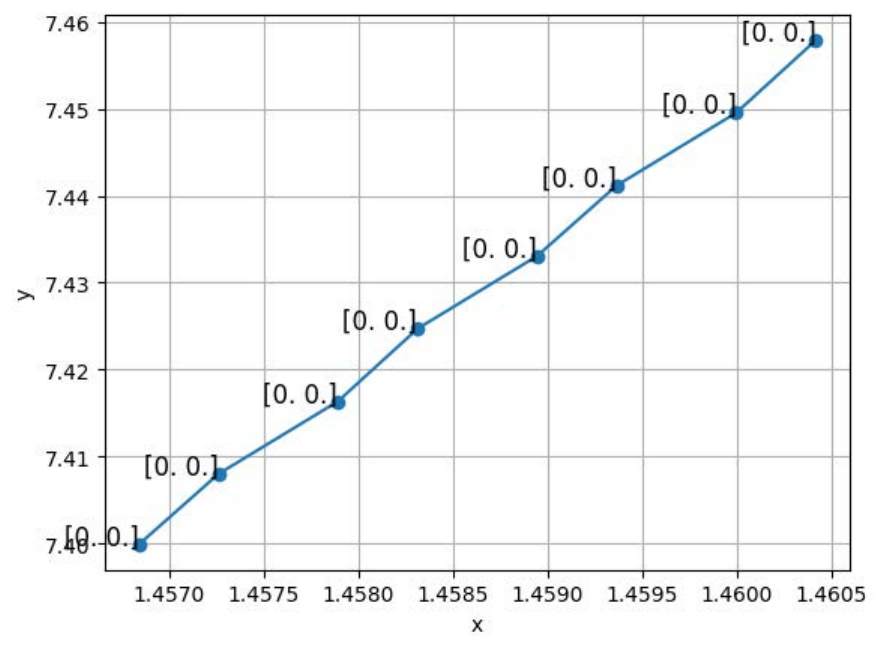}
    \caption{Intention label of the trajectory of a pedestrian}
    \label{fig:ma_vis}
\end{figure}

The output of the intention computation is the intention labels for each pedestrian at each frame. Fig. \ref{fig:ma_vis} exhibits the intention labels of a pedestrian. The trajectory deviates from typical linear motion yet remains within the defined threshold, rendering both the lateral and longitudinal labels as '0' throughout the observed duration.

    \begin{algorithm}
    \caption{Calculate Intention  Label}
    \label{alg:ma_compute} 
    \hspace*{\algorithmicindent} \textbf{Input:} $\mathbf{X} = \left\{\left(x_i^t, y_i^t\right) | t \in \mathcal{T}_o, i \in \mathcal{N} \right\}$ \\
    \hspace*{\algorithmicindent} \textbf{Parameter:} $D_{lat}=0.1,D_{lon} = 0.2,\Delta t =0.8$ \\
    \hspace*{\algorithmicindent} \textbf{Output:} $ \mathbf{I_o} = \left\{(\alpha_i^t, \beta_i^t)| t \in \mathcal{T}_o, i \in \mathcal{N} \right\}$ \\ 
    \hspace*{\algorithmicindent}\qquad \qquad$\mathbf{I_p} = \left\{(\alpha_i^t, \beta_i^t)| t \in \mathcal{T}_p, i \in \mathcal{N} \right\}$
    \begin{algorithmic}[1]
    \If{$y'^{(t+\Delta t)} - y'^t < -D_{lat} $}
    \State $\alpha^t_i \gets 2$ \Comment{turn right}
    \ElsIf{$y'^{(t+\Delta t)}_i - y'^{(t)}_i > D_{lat} $}
    \State $\alpha^t_i \gets 1 $  \Comment{turn left}
    \Else 
    \State $\alpha^t_i \gets 0$ \Comment{keep direction}
    \EndIf
    \If{$(x'^{(t+\Delta t)}_i - x'^{(t)}_i)/ \Delta t  >1 + D_{lon} $}
    \State $\beta^t_i \gets 2$ \Comment{acceleration}
    \ElsIf{$(x'^{(t+\Delta t)} - x'^t)/ \Delta t < 1 - D_{lon} $}
    \State $\beta^t_i \gets 1$ \Comment{deceleration}
    \Else 
    \State $\beta^t_i \gets 0$ \Comment{keep speed}
	\EndIf

    \end{algorithmic}
    \end{algorithm}

\subsection{Spatial-temporal (S-T) module}
This block is similar to the social STGCNN \cite{mohamed2020social-social-stgcnn}, including a Spatial Temporal Graph Convolutional Networks (STGCN) \cite{yan2018spatial-stgcn} block and a 2D CNN with 1D CNN as residual, which acts as a residual architecture. We use the relative position as the input of the spatial block. The input of the spatial is 
$    \mathbf{\Bar{X}} = \left[\mathbf{\Bar{x}}^{T_1}, \mathbf{\Bar{x}}^{T_2}, \ldots, \mathbf{\Bar{x}}^{T_{o}} \right]$,where $    \mathbf{\Bar{x}}^{t} = \left[0,0,x_2^t-x_1^t, y_2^t-y_1^t, \ldots, x_{N}^t-x_{N-1}^t, y_{N}^t - y_{N-1}^t\right]$
we define graphs $G^t$ based on the distance of pedestrians at every time step. $G^t=\left(V^t,E^t,F^t\right)$, where $V^t = \left\{v^t_i|i\in \left\{1,2,\ldots, N\right\} \right\}$ is the set of nodes including the pedestrians at time t, $E^t = \left\{e^t_{ij}|\forall i,j\in \left\{1,2,\ldots, N\right\} \right\}$ is the set of edges indicating the distance of every pedestrian, the $F^t = \left\{f^t_i|i\in \left\{1,2,\ldots, N\right\} \right\} \in \mathbb{R}^{D_f \times N}$ is the feature of every node.$D_f$ is the feature dimension. For the first layer of the STGCN, the feature is the coordinates of every pedestrian. Besides, the adjacent matrix at time $t$ is $A^t$. 
At every time step, $G_t$ is processed by GCN layer \cite{kipf2016semi}. $A= \left\{A^1,A^2,\ldots,A^{T_o}\right\}$. To consider the self-influence, we add the 
identity matrix to the $A^t$ \cite{xu2022adaptive}.
\begin{equation}
    \hat{A^t} = A^t + I
\end{equation}
Then the $\hat{A} = \left\{A^{T_1},\ldots,A^{T_O}\right\}$. Similarly, the $F^{(l)} = \left\{F^{T_1},\ldots,F^{T_o}\right\}$. $F^{(l)} \in \mathbb{R}^{D_f\times  L_o\times N}$ is the feature matrix at $lth$ GCN layer. When $l=0$, the feature is the position of every pedestrian. Besides, the node degree matrices $D = \left\{D^{T_1},\ldots, D^{T_o}\right\}$. The degree matrix can also be derived from the adjacency matrix. For instance, for node $v$, its degree $D^t_{v}$ would equal to $\sum_{u}{A_{uv}}$, which is equal value to the number of all elements in row $v$ of A. The degree matrix is diagonal. The $(l+1)th$ GCN layer is calculated as follows:

\begin{equation}
    F^{(l+1)} = \sigma \left( D^{-\frac{1}{2}} \hat{A} D^{-\frac{1}{2}} F^{(l)} W^{(l)} \right)
\end{equation}
where $\hat{A}=\left\{\hat{A^{T_1}},\ldots,\hat{A^{T_o}}\right\}$,$W^{(l)}$ is are learnable parameters of the $lth$ GCN layer, $\sigma$ is nonlinear function.

\begin{figure}
    \centering
    \includegraphics[width=0.6\linewidth]{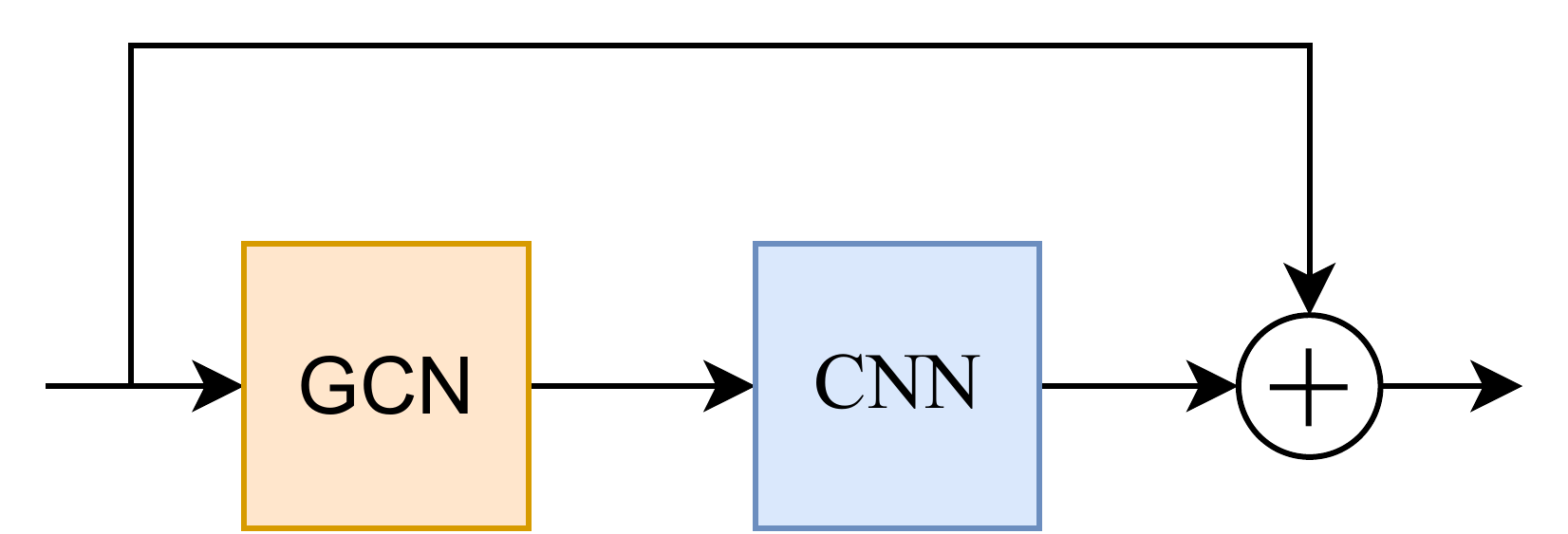}
    \caption{STGCN block}
    \label{fig:stgcn}
\end{figure}

After a GCN layer produces the feature $F^{\left(l\right)}$, it is then inputted into CNN layer. The kernel of the CNN layer is $3 \times 1$, indicating that the CNN layer captures information over three consecutive time steps.
The STGCN block is depicted in Fig. \ref{fig:stgcn} showing that the input of the STGCN is initially processed by a GCN layer to capture spatial information, and then it is passed into a CNN layer for modeling temporal information. After several STGCN blocks, a 2D CNN layer and a residual CNN layer follow.
The output of the S-T module is $F_{ST} \in \mathbb{R}^{D_f\times  L_o\times N}$, which is the spatial-temporal representation of all pedestrians during the observed over time.
\subsection{Temporal module}

The future intention representation is denoted as $F_{I} \in \mathbb{R}^{D_f\times L_o\times N}$, containing the intention information of each pedestrian in the predicted time. The learnable weights $W_{I}$ determine how much intention information is inputted into the temporal CNN. The input for the temporal CNN is $\left(F_{I}*W_I + F_{ST}\right) \in \mathbb{R}^{D_f\times L_o\times N}$. The output of the temporal CNN is the predicted trajectory $\hat{Y} \in \mathbb{R}^{D_f\times L_p\times N}$. The temporal CNN is a 2D CNN layer that transforms the channel from $L_o$ to $L_p$ and produces the predicted trajectory.

\subsection{CRF module}
Inspired by MSDC \cite{he2023msdc}, using CRF, a probabilistic graphical model, to capture more information in the input data of power consumption by adding CRF loss to add constraints on the state label. This work incorporates CRFs into the trajectory prediction model, allowing for the capture of complex dependencies between different intentions. 
The use of CRFs aligns with the theory that pedestrians' movements are not isolated events but are influenced by surrounding pedestrians and historical data. 

In trajectory prediction, CRFs model the conditional probability of a sequence of intention given a sequence of 
observations. The input of CRF is an emission feature vector $E\in \mathbb{R}^{B\times L\times N_g}$, which is a value for each possible intention state. Where the $B$ is the batch size, $L$ is the length of the sequence, and $N_g$ is the number of ground truth labels. This value represents the probability or score of the intention label in the predicted trajectory data. However, the intention label corresponding to the real situation is not always the biggest value. Because the intention of the 
predicted time has constraints at the previous time. For example, if the pedestrian is turning left at the previous time, the pedestrian can not turn left again at the current time. So, the CRF is used to capture the constraints between the two intention labels. 
\subsubsection{Time CRF Module}
\begin{figure}
    \centering
    \includegraphics[width=0.9\linewidth]{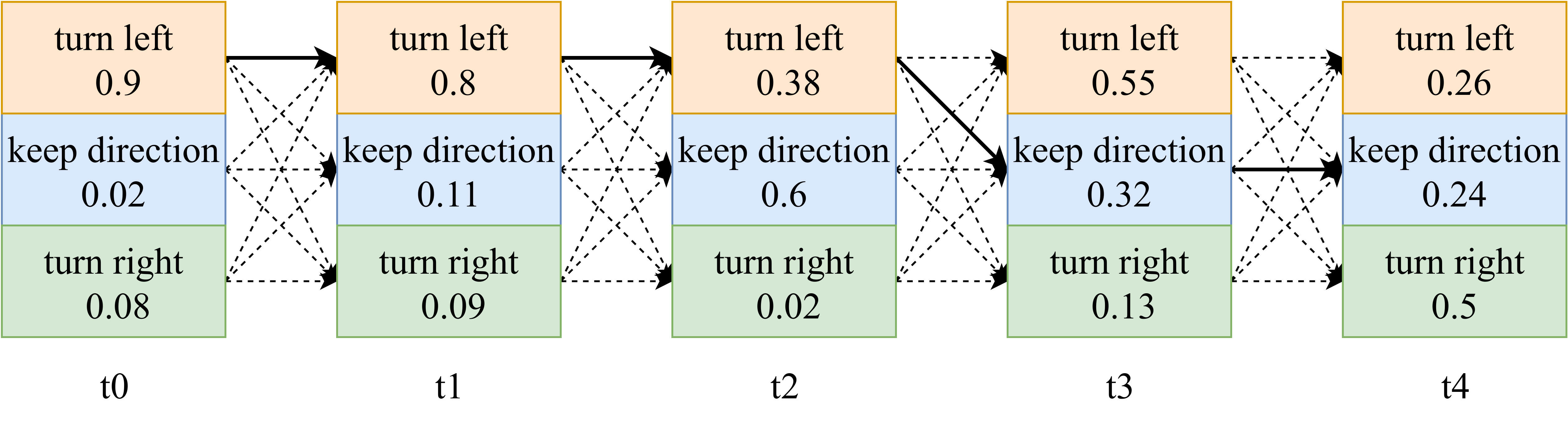}
    \caption{Emission score of lateral Time CRF}
    \label{fig:emission_timecrf_lat}
\end{figure}

\begin{figure}
    \centering
    \includegraphics[width=0.9\linewidth]{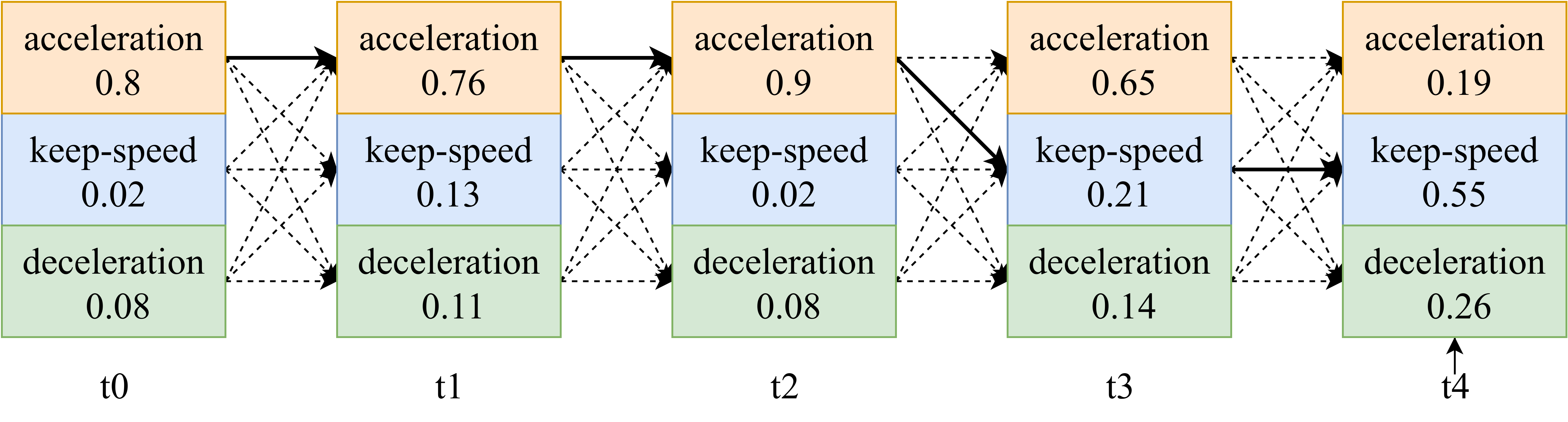}
    \caption{Emission score of longitudinal Time CRF}
    \label{fig:emission_timecrf_lon}
\end{figure}

\textbf{Time CRF:} The output of S-T module, $F_{ST} \in \mathbb{R}^{D_f\times  L_o\times N}$ is processed by linear layer to obtain the emission matrix, including the emission score of every pedestrian of Time CRF, which is denoted as $E_T  \in \mathbb{R}^{N\times L_p\times N_I}$. $N_I = 3$ is the number of intention labels, which is three both for longitudinal and lateral intention. The $E_T$ represents the likelihood of every intention label at every time step.
As Fig.\ref{fig:emission_timecrf_lat} and Fig. \ref{fig:emission_timecrf_lon} show, the emission score of the Time CRF is the most probability of every intention of the pedestrian at every time step. Each box in Fig. \ref{fig:emission_timecrf_lat} represents the highest likelihood of a pedestrian's lateral intention at different frames, while the boxes in Fig. \ref{fig:emission_timecrf_lon} represent the highest likelihood of a pedestrian's longitudinal intention at different frames. In Fig. \ref{fig:emission_timecrf_lat}, the intention sequence will be [turn left, turn left, keep direction, turn left, turn right], which is different from the real sequence [turn left, turn left, turn left, keep direction, keep direction]. Taking into account the intention constraints in the adjacent frame, the true intention sequence does not select the most probable intention at each frame. The CRF is an excellent tool for calculating these constraints and can generate the accurate intention sequence as output by computing the transition score $\Psi$. CRF is applied to capture these constraints, which will make the intention more reasonable. Through the Time CRF, the predicted intention $\hat{I}_p \in \mathbb{R}^{D_f\times L_p\times N}$ is computed by the Viterbi algorithm \cite{sutton2012introduction-crf}. Then, a CNN layer is used to get the future intention representation $F_{I} \in \mathbb{R}^{D_f\times L_p\times N}$, which includes the intention information for every pedestrian in the prediction time. The other output of the Time CRF is the
$\mathcal{L}_T = \mathcal{L}_T^{lat} + \mathcal{L}_T^{lon} $, which is computed based on the 
true intention of every pedestrian in the prediction time shown in Eq. \ref{eq:time_crf_loss}. 
\begin{equation}
    \begin{aligned}
        \mathcal{L}_T^{lat} &= - \mathbb{E} \left[ \sum_{t=T_o+1}^{T_p} e_{t,\alpha_i^t} + \sum_{t=T_o+1}^{T_p-1} \Psi(\alpha_i^t, \alpha_i^{t+1}) \right] 
        + \log Z_T^{lat}\\
        \mathcal{L}_T^{lon} &= - \mathbb{E} \left[ \sum_{t=T_o+1}^{T_p} e_{t,\beta_i^t} + \sum_{t=T_o+1}^{T_p-1} \Psi(\beta_i^t, \beta_i^{t+1}) \right] 
        + \log Z_T^{lon}
    \end{aligned}
    \label{eq:time_crf_loss}
\end{equation}

Where $\sum_{t=T_o+1}^{T_p} e_{t,\alpha_i^t}$ and $\sum_{t=T_o+1}^{T_p} e_{t,\beta_i^t}$ represent the sum of probabilities for all predicted intentions, with the former being the lateral emission score and the latter being the longitudinal emission score.

The transition score,$\Psi(\alpha_i^t, \alpha_i^{t+1})$,$\Psi(\beta_i^t, \beta_i^{t+1})$ are computed from all predicted intention distributions in the prediction time. $\log Z_T^{lat}$ and $\log Z_T^{lon}$ ensure the probability distribution is normalized so that the sum of probabilities over all possible state sequences equals 1.
\subsubsection{Space CRF Module}
\textbf{Space CRF:} The space CRF focuses on capturing the interaction between pedestrians based on intention labels. After the linear layer, $F_{ST}$ can be transformed into the emission matrix of Space CRF, denoted as $E_S$. Here, $F_{ST}$ is in $\mathbb{R}^{D_f\times  L_o\times N}$ and $E_S$ is in $\mathbb{R}^{L_o\times N\times  N_I}$. In the Space CRF, the batch size of the emission score corresponds to the trajectory's time length, while the number of pedestrians represents the sequence length.
This shape of emission enables Space CRF to process the interaction between pedestrians. 
The output of Space CRF is only the space CRF loss $\mathcal{L}_{S}$, which is computed based on the true intention of every pedestrian in the observations time. 
This loss function ensures that the spatial relationships between pedestrians are accurately represented, enhancing the model's ability to predict complex interactions. The mathematical formulation of Space CRF loss can be expressed as:
\begin{equation}
    \begin{aligned}
        \mathcal{L}_S^{lat} &= - \mathbb{E} \left[ \sum_{t=T_2}^{T_o} e_{t,\alpha_i^t} + \sum_{t=T_1}^{T_o-1} \Psi(\alpha_i^t, \alpha_i^{t+1}) \right] 
        + \log Z_S^{lat}\\
        \mathcal{L}_S^{lon} &= - \mathbb{E} \left[ \sum_{t=T_2}^{T_o} e_{t,\beta_i^t} + \sum_{t=T_1}^{T_o-1} \Psi(\beta_i^t, \beta_i^{t+1}) \right] 
        + \log Z_S^{lon}
    \end{aligned}
    \label{eq:space_crf_loss}
\end{equation}
Where $\sum_{t=T_2}^{T_o} e_{t,\alpha_i^t}$ and $\sum_{t=T_2}^{T_o} e_{t,\beta_i^t}$ represent the sum of probabilities for all observed intentions. $\Psi(\alpha_i^t, \alpha_i^{t+1})$,$\Psi(\beta_i^t, \beta_i^{t+1})$ are the transition score. $\log Z_S^{lat}$ and $\log Z_S^{lon}$ ensures the probability distribution is normalized so that the sum of probabilities over all possible state sequences equals 1.

With the space CRF loss, the interaction among pedestrians is captured based on their movement intention. The loss enhances the S-T module, which only captures the interaction based on the representation of pedestrians' position. The space CRF loss benefits the S-T module by capturing interactions and making the interaction representation more explainable, transforming it from a black box to a grey box.

So, the total Loss function of the ST-CRF model is like the following:

\begin{equation}
    \mathcal{L} = \mathcal{L}_{S} + \mathcal{L}_{T} +||Y-\hat{Y}||_1
\end{equation}
\section{Experiments}
\label{sec:experiment}

\subsection{Datasets}
To evaluate our model, we perform experiments on the ETH/UCY dataset, which includes five scenes: ETH, Hotel, Univ, Zara1, and Zara2. In the ETH/UCY dataset, the observation time is 8 frames, and the prediction time is 12 frames. Each frame has a duration of 0.4 seconds. Additionally, we conduct experiments on the SDD. We follow the Social Implicit approach \cite{mohamed2022social-Social-implicit} for setting up these datasets.

\subsection{Evaluation Metrics}
The performance of the model is evaluated using three key metrics, each serving a specific purpose in assessing different aspects of the prediction accuracy:
\begin{itemize}
    \item \textbf{Average Displacement Error (ADE):} Defined in Equation \ref{eq:ade}, ADE measures the average Euclidean distance between the predicted positions $\hat{y}_{n}^t$ and the ground truth positions $y_{n}^t$ overall predicted time steps $t$ in $T_{pred}$ and all vehicles $n$ in $N$. This metric quantifies the overall deviation of the predicted trajectories from the actual paths, providing a general assessment of the model's accuracy.
    \begin{equation}
        ADE=\frac{\sum _{ t \in \mathcal{T}_p }\sum_{i \in \mathcal{N}}\|\hat{y}_{i}^t-y_{i}^t\|_2}{N}
        \label{eq:ade}
    \end{equation}
    \item \textbf{Final Displacement Error (FDE):} As shown in Equation \ref{eq:fde}, FDE calculates the Euclidean distance between the predicted and actual positions at the final time step $T_{pred}$ for all vehicles $n$ in $N$. This metric specifically evaluates the model's ability to accurately predict the endpoint of the trajectories, which is crucial for long-term prediction tasks.
    \begin{equation}
        FDE=\frac{\sum_{i \in \mathcal{N}}\|\hat{y}_{i}^{T_{p}}-y_{n}^{T_{p}}\|_2}{N}
        \label{eq:fde}
    \end{equation}
    \end{itemize}
\begin{table*}[]
\centering
\caption{Prediction performance on ETH and UCY dataset. The lower the better. The bold font represents the best result}
\label{tab:eth-ucy}
\begin{tabular}{cccccccc}
 \toprule
 & \multicolumn{6}{c}{Performance(ADE/FDE)} \\
 \midrule
 Model&Venues & ETH & Hotel & Univ & Zara1 & Zara2 &Average \\
\midrule
S-LSTM \cite{alahiSocialLSTMHuman2016a-sociallstm}& CVPR'2016 &1.09/2.35 & 0.79/1.76& 0.67/1.40 &0.47/1.00& 0.56/1.17& 0.72/1.54 \\
 S-GAN \cite{gupta2018social-socialgan}& CVPR'2018 &0.81/1.52 & 0.72/1.61 & 0.60/1.26  &  0.34/0.69  & 0.42/0.84     &  0.58/1.18    \\
S-STGCNN \cite{mohamed2020social-social-stgcnn} &CVPR'2020& 0.64/\textbf{1.11} & 0.49/0.85 & 0.44/0.79  & 0.34/0.53  & 0.30/0.48 & 0.44/0.75   \\
 CARP$^e$ \cite{Mendieta_Tabkhi_2022-CARPe}&AAAI'2021 &0.80/1.48 &0.52/1.00&0.61/1.23& 0.42/0.84 & 0.34/0.69 & 0.54/1.05\\
 Social-Implicit \cite{mohamed2022social-Social-implicit}&ECCV'2022& \underline{0.66}/1.44 & \underline{0.20}/\underline{0.36} & \underline{0.31}/\underline{0.60} & \underline{0.25}/\underline{0.50} & \underline{0.22}/\underline{0.43} & \underline{0.33}/\underline{0.67}\\
  Pishgu \cite{alinezhad2023pishgu} & ICCPS'2023 & 1.10/2.24 & \textbf{1.17}/2.17 & 0.67/1.37 & 0.45/0.91 & 0.36/0.73& 0.75/1.48\\
 \midrule
 \textbf{ST-CRF (ours) }& -&\textbf{0.61}/\underline{1.32} & \textbf{0.17/0.28} &	\textbf{0.29/0.55} &	\textbf{0.24/0.46} & \textbf{0.21}/\textbf{0.42} & \textbf{0.30}/\textbf{0.60}\\
 \bottomrule
\end{tabular}
\end{table*}

\subsection{Quantitative Analysis}
The results of the experiments on the ETH/UCY dataset are summarized in Table \ref{tab:eth-ucy}. The evaluation metrics used are ADE and FDE. In terms of both ADE and FDE, the ST-CRF method consistently outperforms other methods in Hotel, Univ, Zara1, and Zara2. For ETH, only the FDE is larger than that of S-STGCNN. This implies that the ST-CRF provides more accurate predictions for the future trajectory at the final time step and short-term trajectory.The ST-CRF backbone is similar to Social-Implicit, with the addition of intention computer and space CRF, as well as time CRF. Our work, ST-CRF, achieves better results than Social-Implicit for all datasets in ETH/UCY. For average ADE and FDE on ETH/UCY, ST-CRF is 9.09\% better in average ADE and 10.44\% better in average FDE. The results indicate that these three modules significantly enhance performance.

In addition, we conduct experiments on SDD. Table \ref{tab:sdd_result} shows that the ST-CRF model achieves SOTA. For ADE, ST-CRF is 10.64\% better than the SOTA model Social-Implicit and 14.61\% better than another SOTA model Social-Implicit.

\begin{table}[H]
\centering
\caption{The performance on the SDD dataset.  The lower, the better.}
\label{tab:sdd_result}
\begin{tabular}{l|l|ll}
\toprule
        Model &Venues& ADE  & FDE  \\
\midrule
STGAT \cite{huang2019stgat}&ICCV'2019& 0.58 & 1.11 \\
Social-Ways \cite{amirian2019social-socialways}&CVPR'2019& 0.62 & 1.16 \\
DAG-Net \cite{monti2021dag-dagnet}&ICPR'2021& 0.53 & 1.04 \\
Social-Implicit\cite{mohamed2022social-Social-implicit} &ECCV'2022& \underline{0.47} & \underline{0.89} \\
\midrule
\textbf{ST-CRF (ours) } & & \textbf{0.42} & \textbf{0.76}\\
\bottomrule
\end{tabular}
\end{table}



\subsection{Alation study}
We performed ablation experiments on ETH/UCY to investigate the impact of space and time CRF loss. Table \ref{tab: ablation} demonstrates that removing all CRF loss yields the poorest result. Using only $\mathcal{L}_S$ or $\mathcal{L}_T$ produces intermediate performance, but it is still inferior to using both $\mathcal{L}_S$ and $\mathcal{L}_T$.

\begin{table}[]
\caption{Ablation study on ETH/UCY dataset}
\centering
\label{tab: ablation}
\begin{tabular}{cc|cc}
\toprule
$\mathcal{L}_T$                               & $\mathcal{L}_S$           & \begin{tabular}[c]{@{}c@{}}Average \\ ADE\end{tabular} & \begin{tabular}[c]{@{}c@{}}Average \\ FDE\end{tabular} \\
\midrule
     & \checkmark &    0.31             &  0.63                                                     \\
\checkmark &      &       0.32           &    0.63                     \\
\checkmark                     & \checkmark &  \textbf{0.30 }          & \textbf{0.60}\\ 
\bottomrule                                                  
\end{tabular}
\end{table}

\begin{table}[]
\caption{The performance of S-STGCNN improved with the addition of time CRF loss and space CRF loss}
\label{tab:s-stgcnn add}
\begin{tabular}{c|cc|cc}
\toprule
Model                                                             & $\mathcal{L}_T$           & $\mathcal{L}_S$           & \begin{tabular}[c]{@{}c@{}}Average \\ ADE\end{tabular}                                    & \begin{tabular}[c]{@{}c@{}}Average \\ FDE\end{tabular} \\ 
\midrule
S-STGCNN \cite{mohamed2020social-social-stgcnn}                                                          &                           &                           & 0.44                                                                                      & 0.75                                                                                   \\
\midrule
\begin{tabular}[c]{@{}c@{}}S-STGCNN \\ (new version)\end{tabular} & \checkmark & \checkmark & \begin{tabular}[c]{@{}c@{}}0.38\\ $\left(\downarrow \textbf{13.63\%}\right)$\end{tabular} & \begin{tabular}[c]{@{}c@{}}0.66\\ $\left(\downarrow \textbf{12.00\%}\right)$\end{tabular}\\ \bottomrule
\end{tabular} 
\end{table}

Time CRF loss and space CRF loss were added to S-STGCNN, resulting in improved performance, as shown in Table 1.   The average ADE decreased by 13.63\% and the average FDE decreased by 12.00\%. The results demonstrate the improvement in performance of the previous deep learning model due to the addition of these two losses to S-STGCNN. In addition, the inclusion of the intention computer can be plug-and-play, making the addition of the two losses to S-STGCNN very easy.

\begin{figure}
    \centering
    \begin{subfigure}[b]{0.22\textwidth}
        \centering
        \includegraphics[width=0.85\textwidth]{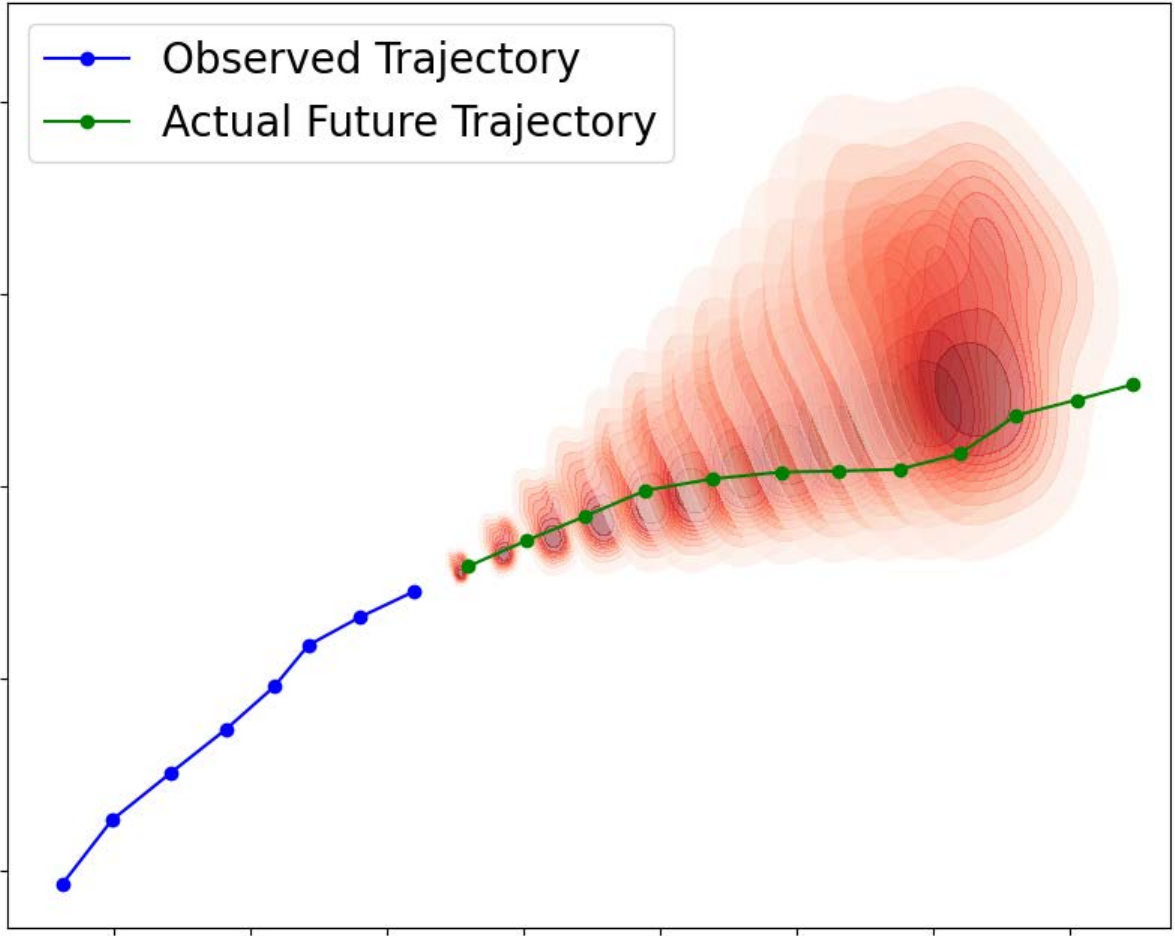}
        \caption{ST-CRF}
        \label{fig:stcrf-dist}
    \end{subfigure}
    \begin{subfigure}[b]{0.22\textwidth}
        \centering
        \includegraphics[width=0.85\textwidth]{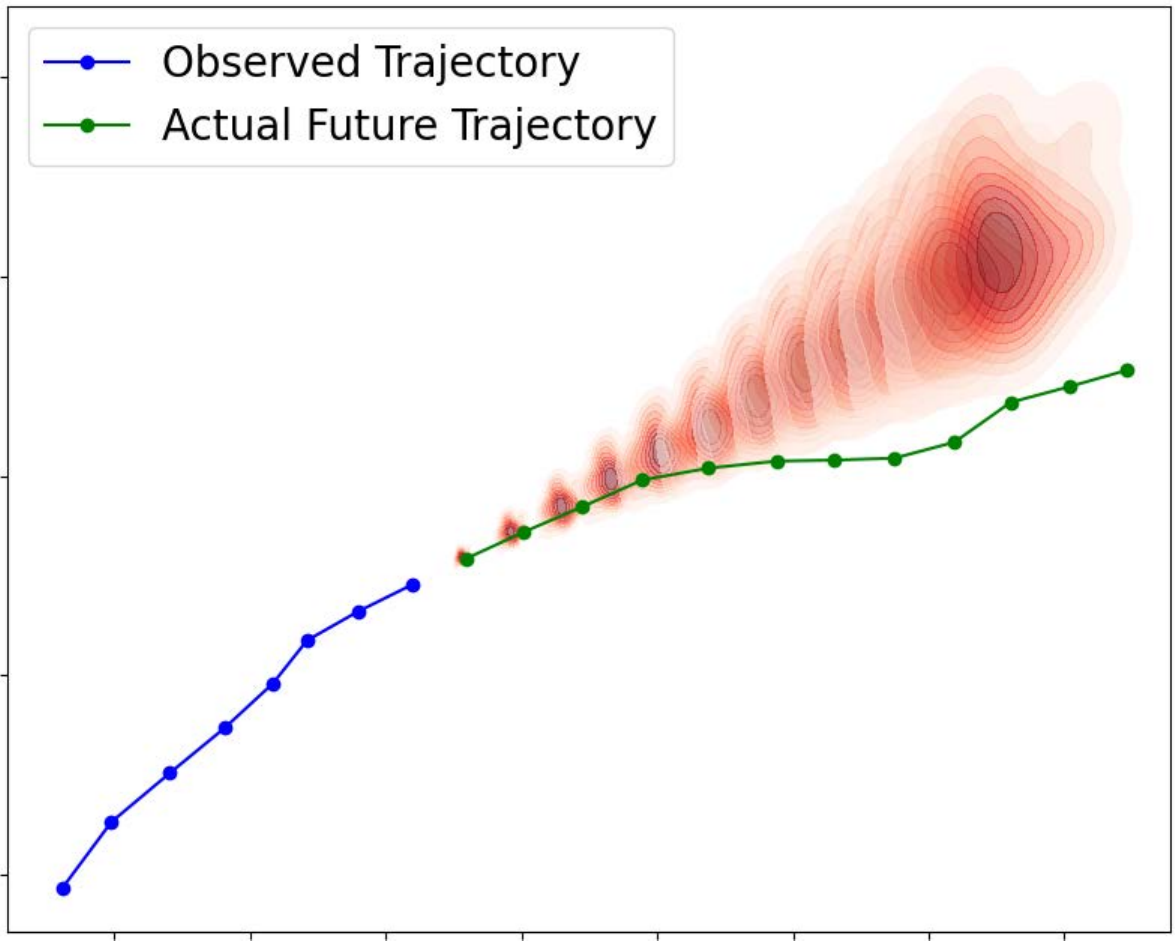}
        \caption{Social-Implicit}
        \label{fig:si-dist}
    \end{subfigure}
    
    \caption{Distribution of Social-Implicit and ST-CRF}
    \label{fig:distribution} 
\end{figure}
Fig. \ref{fig:distribution} presents a case study on the distribution of future trajectories. It demonstrates that the distribution of future trajectories is significantly superior to that of the Social-Implicit.
\subsection{ Implementation Details}
We trained our model on a Linux server running Ubuntu 22.04.1 LTS, equipped with an AMD Ryzen Threadripper PRO 3995WX processor (64 cores), 512GB RAM, and one NVIDIA GeForce RTX 3090 GPU. The CRF loss is implemented based on a Pytorch library. There is a question regarding the selection of the time interval $\Delta t$. We discovered that each frame in the ETH/UCY dataset lasts 0.4 seconds. Additionally, according to \cite{laplante2004continuing-ped-speed}, pedestrians have an average speed of 1.2 m/s. Therefore, we set the time interval $\Delta t$ to 0.8 seconds, resulting in intention labels being calculated over a distance of 0.96 meters, which is a reasonable distance to compute the intention labels. We use two STGCN blocks in the S-T module.

\section{Conclusion}

In the paper, we have developed a novel model called ST-CRF that takes into account pedestrians' intentions to predict their future trajectories. This model can incorporate spatial, temporal, and intention information from trajectory data. We conducted experiments on ETH/UCY and SDD datasets, which demonstrate that ST-CRF outperforms the baseline model. These results provide evidence for our claims. Additionally, the ablation study reveals that both CRF loss and intermediate variable (future intention representation of pedestrians) contribute to accurate trajectory prediction. Furthermore, the intention computer and CRF loss are modular components that can be easily integrated into various deep learning models for trajectory prediction purposes.
{
    \small
    \bibliographystyle{ieeenat_fullname}
    \bibliography{main}
}


\end{document}